%% file: ecai.tex
\documentclass{ecai} 
\usepackage{graphicx}
\usepackage{latexsym}

\usepackage{color}
\usepackage{algorithm}
\usepackage{algorithmic}

\definecolor{bluegray}{rgb}{0.4, 0.6, 0.8}
\definecolor{taupegray}{rgb}{0.55, 0.52, 0.54}
\definecolor{steelblue}{rgb}{0.27, 0.51, 0.71}

\usepackage{hyperref} 
\hypersetup{
  colorlinks   = true, 
  urlcolor     = blue, 
  linkcolor    = blue, 
  citecolor   = blue 
}
\usepackage[labelsep=period]{caption}
\DeclareCaptionFormat{centered}{\captsize\centering #1#2#3\par}
\usepackage{inconsolata}
\usepackage{amsmath}
\DeclareMathOperator*{\argmax}{arg\,max}
\usepackage{microtype}
\usepackage{subfigure}
\usepackage{booktabs} 
\usepackage{nicematrix}
\usepackage{stfloats}

\usepackage{xcolor,colortbl}
\definecolor{green}{rgb}{0.1,0.1,0.1}

\definecolor{mycolor}{rgb}{1.0,0.7,0.1}

\ecaisubmission   

\begin{document}

\begin{frontmatter}

\title{Shrink-Perturb Improves Architecture Mixing during Population Based Training for Neural Architecture Search}

\author[A]{\fnms{Alexander}~\snm{Chebykin}\thanks{Corresponding Author. Email: a.chebykin@cwi.nl.}\orcid{0000-0002-3549-3533}}
\author[A]{\fnms{Arkadiy}~\snm{Dushatskiy}\orcid{0000-0003-0945-0262}}
\author[B]{\fnms{Tanja}~\snm{Alderliesten}\orcid{0000-0003-4261-7511}}
\author[A, C]{\fnms{Peter}~\snm{Bosman}\orcid{0000-0002-4186-6666}}

\address[A]{Centrum Wiskunde \& Informatica}
\address[B]{Leiden University Medical Center, Department of Radiation Oncology}
\address[C]{Delft University of Technology}

\input{0_abstract}

\end{frontmatter}

\input{1_intro}

\input{2_0_related}

\input{3_method}

\input{4_experiments}

\input{5_discussion}

\input{6_conclusion}

\ack This work is part of the research projects DAEDALUS (funded via the Open Technology Programme of the Dutch Research Council (NWO), project number 18373; part of the funding is provided by Elekta and ORTEC LogiqCare) and OPTIMAL (funded by NWO, project OCENW.GROOT.2019.015). 

\bibliography{ecai}

\input{7_appendix}

\end{document}

%% file: 0_abstract.tex
\begin{abstract}

In this work, we show that simultaneously training and mixing neural networks is a promising way to conduct Neural Architecture Search (NAS).  
For hyperparameter optimization, reusing the partially trained weights allows for efficient search, as was previously demonstrated by the Population Based Training (PBT) algorithm.
We propose PBT-NAS, an adaptation of PBT to NAS where architectures are improved during training by replacing poorly-performing networks in a population with the result of mixing well-performing ones and inheriting the weights using the shrink-perturb technique. After PBT-NAS terminates, the created networks can be directly used without retraining. PBT-NAS is highly parallelizable and effective: on challenging tasks (image generation and reinforcement learning) PBT-NAS achieves superior performance compared to baselines (random search and mutation-based PBT).

\end{abstract}

%% file: 1_intro.tex
\section{Introduction}
\label{intro}

Neural Architecture Search (NAS) is the process of automatically finding a neural network architecture that performs well on a target task (such as image classification~\cite{liu2018darts}, natural language processing~\cite{klyuchnikov2022bench}, image generation~\cite{gao2020adversarialnas}). One of the key questions for NAS is the question of efficiency, since evaluating every promising architecture by fully training it would require an extremely large amount of computational resources.

Many approaches have been proposed for increasing the search efficiency: low-fidelity evaluation~\cite{zoph2016neural, shim2021core}, using weight sharing via a supernetwork~\cite{pham2018efficient, cai2019once}, estimating architecture quality via training-free metrics~\cite{mellor2021neural, abdelfattah2021zero}.
Typically, each approach has two stages: first, finding an architecture efficiently, then, training it (or its scaled-up version) from scratch. 
This final training usually requires a manual intervention (e.g., if an architecture of a cell is searched, determining how many of these cells should be used), which diminishes the benefit of an automatic approach (potentially, this could also be automated, but we are not aware of such studies in the literature). Ideally, an architecture itself (not its proxy version) should be searched on the target problem, with the search result being immediately usable after the search (such single-stage NAS approaches exist but are limited: e.g., they restrict potential search spaces~\cite{hu2020dsnas} or require costly pretraining~\cite{cai2019once, wang2021attentivenas}).

For the task of hyperparameter optimization (which is closely related to NAS), effective and efficient single-stage algorithms exist in the form of Population Based Training (PBT)~\cite{jaderberg2017population} and its extensions~\cite{liang2021regularized, dalibard2021faster}. The key idea of PBT is to train many networks with different hyperparameters (a population) in parallel: as the training progresses, worse networks are replaced by copies of better ones (including the weights), with hyperparameter values explored via random perturbation. PBT is highly efficient due to the weight reuse, and due to its parallel nature: given a sufficient amount of computational resources, running PBT takes approximately the same wall-clock time as training just one network.

Determining the best way to adapt PBT to NAS is an open research question~\cite{dalibard2021faster}: if a network architecture has been perturbed, the partly-trained weights cannot be reused (because, e.g., weights of a convolutional layer cannot be used in a linear one). The naive approach of initializing them randomly does not work well (see Section~\ref{abl:shpe_crucial}), and existing algorithms extending PBT to NAS~\cite{franke2020sample, wan2022bayesian} sidestep the issue at the cost of parallelizability or performance (see Section~\ref{rel:hptune}).

\begin{figure*}[ht]
\centerline{\includegraphics[width=0.75\textwidth]{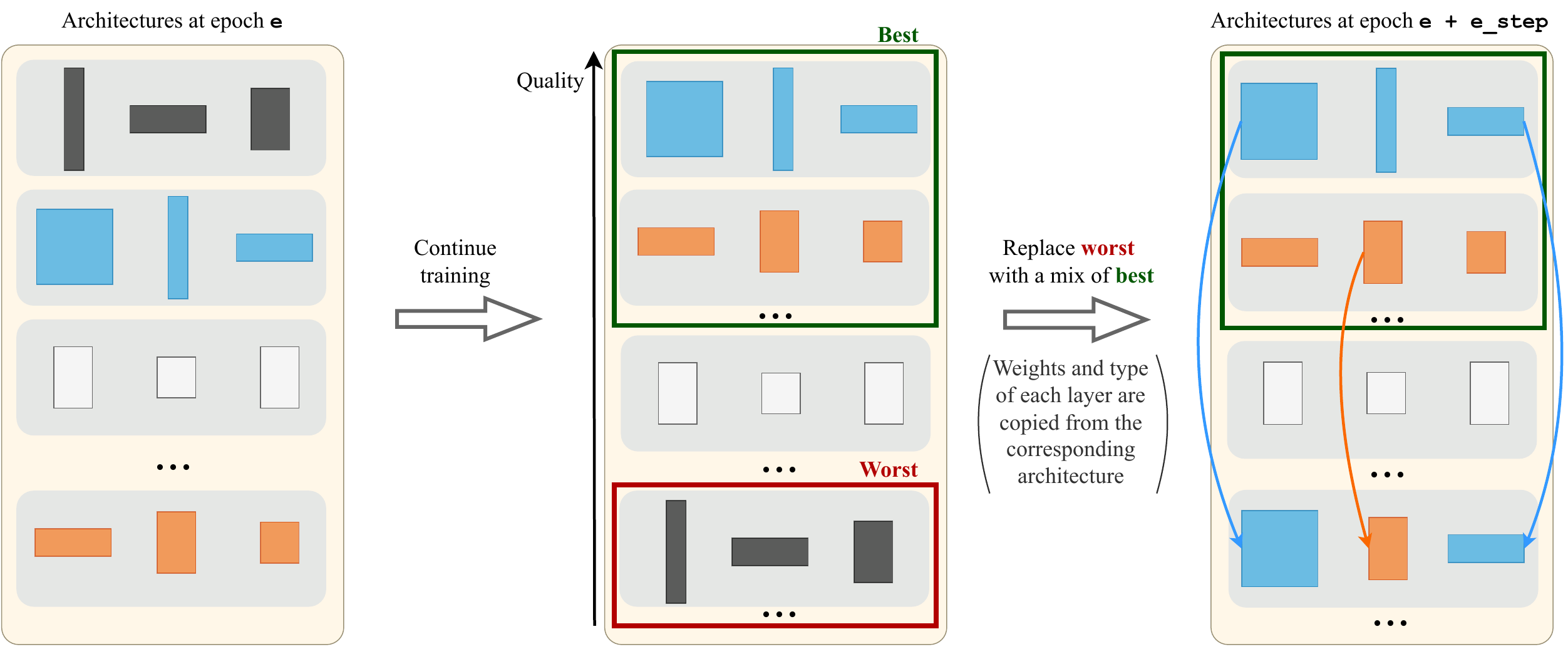}}
\caption{In each iteration of PBT-NAS, architectures in the population continue training for several epochs and then are sorted by performance. Every architecture from the bottom percentile is replaced with a mix of two architectures from the top percentile. During mixing, each layer is copied from one of these two architectures (weights from the architecture with worse performance are shrink-perturbed). Different shapes represent different types of layers.}
\label{fig:algo_step_scheme}
\end{figure*}

We propose to adapt PBT to NAS by modifying the search to rely not on random perturbations but on mixing layers of the networks in the population. An example of this principle is combining an encoder and a decoder from two different autoencoder networks, ultimately obtaining a better-performing network. 
In this setting, the source of the weights for the changed layers is natural: they can be copied from the parent networks. 
Furthermore, we explore if additionally adapting the copied weights with the shrink-perturb technique~\cite{ash2020warm} (reducing weight magnitude and adding noise) is helpful for achieving a successful transfer of a layer from one network to another. 

For many standard tasks (such as image classification), single-objective NAS algorithms are  matched by (or show only a small improvement over) the simple baseline of random search~\cite{li2020random, yuevaluating}.
In order to make the potential benefit of PBT-NAS clear, 
experiments in this paper are conducted in two challenging settings: Generative Adversarial Network (GAN) training, and Reinforcement Learning (RL) for visual continuous control. We further advocate for harder tasks and search spaces in Section~\ref{discussion}.

While our approach could potentially be extended to include hyperparameter optimization, this paper is focused on architecture search.

The contributions of this work are threefold:
\begin{enumerate}
    \item We propose to conduct NAS by training a population of different architectures and mixing them on-the-fly to create better ones (while inheriting the weights).
    \item We investigate if applying shrink-perturb~\cite{ash2020warm} to the weights is a superior technique for weight inheritance compared to copying or random reinitialization.
    \item Integrating these ideas, we introduce PBT-NAS, an efficient and general NAS algorithm, and evaluate it on challenging NAS search spaces and tasks. 
\end{enumerate}

%% file: 2_0_related.tex
\section{Related work}
\label{relwork}

\subsection{Neural Architecture Search} \label{rel:nas}

NAS is the automatic process of finding a well-performing neural network architecture for a specific task. Already in early NAS work~\cite{zoph2016neural}, efficiency concerns played a role: candidate architectures were trained for only a few epochs. Similar low-fidelity search methods save compute by using fewer layers~\cite{lu2019nsga}, or only a subset of the data~\cite{shim2021core}. Another way to save compute is by utilizing a training-free metric to perform NAS without any training~\cite{mellor2021neural, abdelfattah2021zero}.

ENAS~\cite{pham2018efficient} introduced the idea of weight sharing: all candidate architectures are viewed as subsets of a supernetwork, with the weights of the common parts reused across the architectures. The final architecture is scaled up and trained from scratch. This approach greatly decreased cost of the search to just several GPU-days. DARTS~\cite{liu2018darts} further increased efficiency by continuously relaxing the problem. Many approaches build upon DARTS by e.g., reducing memory usage~\cite{xu2020pc} or improving performance~\cite{chen2021progressive}.
AdversarialNAS~\cite{gao2020adversarialnas} extends the approach to GAN training, outperforming previous algorithms~\cite{gong2019autogan}.

In OnceForAll~\cite{cai2019once}, a supernetwork is pretrained such that subnetworks would perform well without retraining, in AttentiveNAS~\cite{wang2021attentivenas}
and AlphaNet~\cite{sharma2020alphanet}
performance is further improved. 
These approaches are a good fit for multi-objective NAS (where in contrast to single-objective NAS, multiple architectures with different trade-offs between objectives such as performance and latency are searched). 
However, the costs for the proposed pretraining reach thousands of GPU-hours. Additionally, in any supernetwork approach, the diversity and size of the architectures are restricted by the supernetwork.

Our approach of exchanging layers and weights between different networks is distinct from the supernetwork-based weight sharing. The weights in the supernetwork are constrained to perform well in a variety of subnetworks, while in our approach, after the weights have been copied to the network with a novel architecture, they can be freely adapted to it, independently of what happens to their original version in the parent network.


The general idea of creating new architectures by modifying existing ones and reusing the weights has been explored in NAS approaches~\cite{elsken2018efficient, jin2019auto} relying on network morphisms\cite{chen2015net2net, wei2016network}.
Network morphisms are operators that change the architecture of a neural network without influencing its functionality. Although \cite{elsken2018efficient, jin2019auto} successfully used morphisms, the idea was later challenged~\cite{wen2020autogrow} with experiments  demonstrating that random initialization of new layers is superior to morphisms. Morphisms are different from our work: while they create a new architecture by modifying one existing architecture, we seek to mix two distinct architectures and reuse their weights.

\subsection{Population Based Training} \label{rel:hptune}

In hyperparameter optimization, hyperparameters of neural network training, such as learning rate or weight decay, are optimized. Bayesian optimization algorithms~\cite{hutter2011sequential, falkner2018bohb} 
are commonly used for sequentially evaluating promising hyperparameter configurations. 
Other approaches include Evolutionary Algorithms~\cite{loshchilov2016cma, liang2021regularized}, and random search~\cite{bergstra2012random}, a simple but reasonably good baseline.

In contrast to the approaches that train weights for each hyperparameter configuration from scratch, PBT~\cite{jaderberg2017population} reuses partly-trained weights when exploring hyperparameters (see Section~\ref{intro} for short description and~\cite{jaderberg2017population} for details). 

To the best of our knowledge, two algorithms were proposed for including architecture search into PBT: SEARL~\cite{franke2020sample} and BG-PBT~\cite{wan2022bayesian}. In SEARL, the architecture is modified by mutation, which can add a linear layer, add neurons to an existing layer, change an activation function, or add noise to the weights. We use a SEARL-like mutation as a baseline. 
In BG-PBT, there are multiple generations; in each generation, network architectures are sampled, initialized with random weights, and their training is sped up via distillation from the best network of the previous generation. This approach adds complexity in the form of multiple generations (the number of which must be manually determined) and using distillation (that would require adaptation to each setting, e.g., GAN training). In addition, sequential generations decrease parallelizability. 
Both SEARL and BG-PBT were proposed exclusively for RL tasks, while we construct PBT-NAS to be a general NAS algorithm.

\subsection{Combining several neural networks into one} \label{rel:combine}

Neural networks can be combined in various ways. In evolutionary NAS~\cite{lu2019nsga} where weights are trained from scratch for each considered architecture, crossover is performed between encodings of architectures. Alternatively, there exist methods combining only the weights of networks that have the same architecture~\cite{uriot2020safe, ainsworth2022git}. Naively averaging the weights leads to a large loss in performance~\cite{ainsworth2022git}, which motivated these approaches to align neurons so that they would represent similar features. Averaging weights without alignment is possible if the weights of the networks are closely related. 
The idea of model soups~\cite{wortsman2022model} is to start with a pretrained model, fine-tune it with different sets of hyperparameters, and greedily search which of the fine-tuned models to average.

In our approach, we mix different architectures \emph{together with the weights} during training, in contrast to evolutionary NAS algorithms combining only architecture encodings, and training the weights from scratch. We also avoid the additional complexity of aligning neurons, instead we continue to train the created network, and allow the gradient descent procedure to adapt the neurons to each other (which is facilitated by shrink-perturb~\cite{ash2020warm}, see Section~\ref{method:motivation}).

%% file: 3_method.tex
\section{Method}
\label{method}

\subsection{Problem setting}

The goal of single-objective NAS is to find a network architecture $\alpha^*$ from a search space $\Omega$ that maximizes an objective function $f$ after training the weights $\theta$:
\begin{equation}
\alpha^* = \argmax_{\alpha \in \Omega} f(\alpha; \theta)
\end{equation}
$\Omega$ typically includes network architecture properties such as the number of layers, types of each layer, and its hyperparameters (e.g., convolution size). We will describe an architecture $\alpha$ by $M$ categorical variables $\{x_i\}_{i=0..M-1}$, each taking $l_i$ possible values. 
Note that in the case where more than one architecture is searched for (e.g., generator and discriminator of a GAN), we consider, for simplicity, $\alpha$ to include architecture parameters of all architectures.

\subsection{Algorithm overview} \label{method:naspbt}

In our algorithm, PBT-NAS, we follow the general structure of PBT, where $N$ networks are trained in parallel\footnote{Note that since each network has a different architecture, it has a different training speed, so to avoid biasing the search towards models that require less training time, we use the synchronous variant of PBT.}.
In each iteration of the algorithm, every network is trained for $e\_step$ epochs. Then, each of the worst $\tau\%$ of the networks is replaced by a mix of two networks from the best $\tau\%$ (according to the objective function $f$). Over time, better architectures are created. Mixing architectures during training is the key component of PBT-NAS. In Section~\ref{method:motivation}, we motivate the choice to do NAS by mixing networks. Further details of how we mix architectures are given in Section~\ref{method:combine}.

A visual representation of one iteration of PBT-NAS is shown in Figure~\ref{fig:algo_step_scheme}, and the pseudocode is listed in Algorithm~\ref{alg:naspbt}.

\begin{algorithm}[tb]
   \caption{PBT-NAS}
   \label{alg:naspbt}
\begin{algorithmic}
    \fontsize{8.65pt}{10.65pt}\selectfont
   \STATE {\bfseries Input:} search space $\Omega$, number of variables $M$, population size $N$, number of epochs $e\_{total}$, step size $e\_{step}$, selection parameter $\tau$, probability $p$ of replacing a layer, parameters $\lambda$, $\gamma$ of shrink-perturb
\end{algorithmic}
\begin{algorithmic}[1]
\fontsize{8.65pt}{10.65pt}\selectfont

   \STATE ${pop} \gets $ \{$N$ random architectures from $\Omega$\}
   
   \STATE ${e} \gets 0$
   \WHILE{${e} < {e\_total}$}
        \FOR[in parallel]{$i \gets 0$ {\bfseries to} $N-1$}
            \STATE train ${pop}_i$ for $e\_step$ epochs
            \STATE ${pop}_i.fitness \gets $  \texttt{evaluate}$({pop}_i)$
        \ENDFOR
        \STATE sort ${pop}$ by $fitness$
        \STATE $best\_nets  \gets $ the best $\tau\%$ nets
        \STATE $worst\_indices \gets $ indices of the worst $\tau\%$ nets
        \FOR{$j$ \textbf{in} $worst\_indices$}
            \STATE $pop_j  \gets $ \texttt{create\_architecture}$(best\_nets, p, M, \lambda, \gamma)$
            
            \COMMENT{the result of mixing, see Algorithm~\ref{alg:combine}}
        \ENDFOR
        \STATE $e \gets e + e\_step$
   \ENDWHILE
\end{algorithmic}
\end{algorithm}

\subsection{Key question when modifying architecture during training: where to get the weights from?} \label{method:motivation}

PBT relies on random perturbations of hyperparameters for exploring the search space while the network weights are being continuously trained. This works well when searching for hyperparameters that can be replaced independently of the weights: e.g., after changing the learning rate, the training can continue with the same weights.

However, searching for an architecture means introducing changes that impact the weights, e.g., changing the type of a layer from linear to convolutional. After such a change, the training process is disturbed: the weights of one type of layer cannot be used in another one.

\begin{figure}[ht]
\centerline{\includegraphics[width=0.95\linewidth]{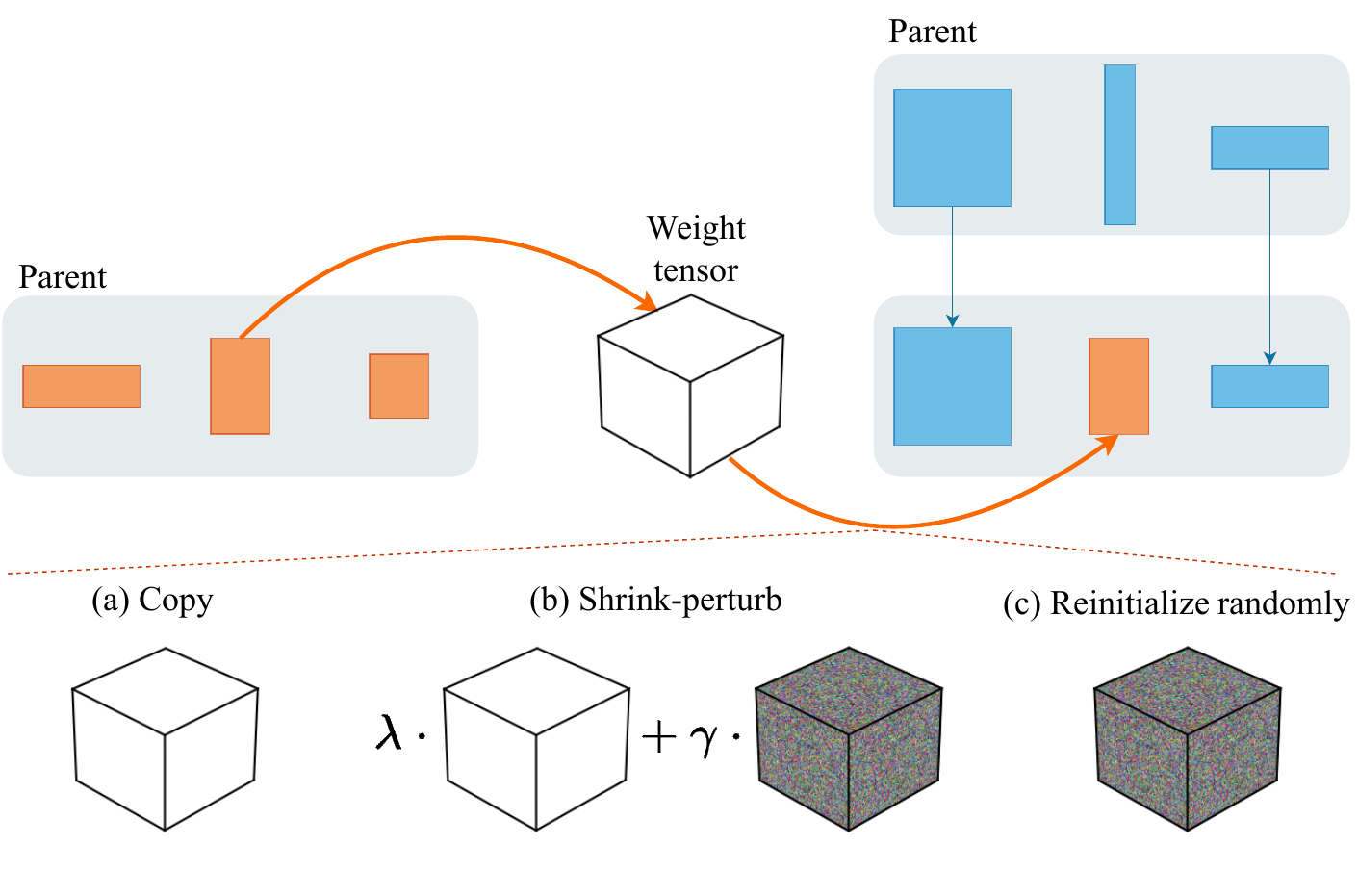}}
\caption{Three potential operations to perform on the weight tensor when copying the corresponding layer from the parent.}    
\label{fig:shrink_perturb_spectrum}
\end{figure}

To follow the paradigm of PBT and continue training the network after an architectural change, the source of the weights needs to be determined. We consider three potential approaches (Figure~\ref{fig:shrink_perturb_spectrum}).

One approach is initializing the new weights randomly. Intuitively, this could be problematic, as replacing weights of a whole layer with random ones can substantially disrupt the learned connections between neurons across the whole network.

Instead of being initialized randomly, the weights of the modified part can come from another network in the population. If the new value of the type of the layer is not generated randomly but copied from another solution, the corresponding layer weights can be copied from it too. 
Straightforward weight copying may be better than random initialization but it faces the following issue: even though the layers at the same depth of different networks should perform similar transformations (when trained on the same task), the actual data representations in each network are likely to be different. The copied weights would need to be adapted to a different representation space, but it might be difficult for gradient descent to adapt them quickly.

\begin{algorithm}[tb]
   \caption{\texttt{create\_architecture}}
   \label{alg:combine}
\begin{algorithmic}
\fontsize{8.65pt}{10.65pt}\selectfont
   \STATE {\bfseries Input:} set of networks to potentially mix $nets$, probability $p$ of replacing a layer, number of variables $M$, parameters $\lambda$, $\gamma$ of shrink-perturb
\end{algorithmic}
\begin{algorithmic}[1]
\fontsize{8.65pt}{10.65pt}\selectfont
    \STATE $net_1, net_2 \gets$ randomly sample from $nets$
   
    \IF{$net_1.fitness < net_2.fitness$}
        \STATE $net_1, net_2 \gets net_2, net_1$ \COMMENT{sort by fitness}
    \ENDIF
    
    \STATE $net_{new} \gets$ \texttt{copy}$(net_1)$
   
    \FOR[iterate over architecture variables]{$i=0$ {\bfseries to} $M-1$}
        \IF{\texttt{random\_uniform}$() < p$}
            \STATE $net_{new}.\alpha_i \gets net_{2}.\alpha_i$ \COMMENT{copy the value of the variable}
            \IF{$\exists net_{2}.W^i$}
                \STATE \COMMENT{if the variable is a layer, copy and modify its weights}
                \STATE $W_{new} \gets$ \texttt{copy}$(net_{2}.W^i)$
                \STATE \texttt{shrink\_perturb}$(W_{new}, \lambda, \gamma)$
                \STATE $net_{new}.W^i \gets W_{new}$
            \ENDIF
        \ENDIF
    \ENDFOR
    
    \STATE \textbf{return} $net_{new}$
    
\end{algorithmic}
\end{algorithm}

Shrink-perturb~\cite{ash2020warm} is potentially helpful in this scenario. It was motivated by the observation that in online learning, continuing training from already trained weights when new data comes in can be worse than retraining from scratch using all the available data. Shrink-perturb consists of modifying the weights of a neural network by \emph{shrinking} (multiplying by a constant $\lambda$) and \emph{perturbing} them (adding noise multiplied by a constant $\gamma$; a new initialization of the network architecture is used as the source of noise). 

Applying shrink-perturb to the copied weights is the middle ground between copying the weights as-is, and initializing them randomly. This preserves some useful information in the weights, while also potentially making their adaptation to the new architecture easier. 

\subsection{Mixing networks} \label{method:combine}

Algorithm~\ref{alg:combine} shows our procedure for creating a new network. Firstly, two parent networks are randomly sampled from the top $\tau$ percentile of the population. An offspring solution is created by copying the better parent, and replacing with probability $p$ each layer with the layer from the worse parent (including the weights, which are shrink-perturbed). Our mixing is a version of uniform crossover~\cite{syswerda1989uniform} where only one offspring solution is produced. 
Note that our mixing requires that layers in the same position can be substituted for each other (i.e., the output can be used as the input of the next layer), with the architecture remaining valid after a layer is replaced. We further discuss this limitation in Section~\ref{discussion}.

Unlike existing approaches to combining neural networks (see Section~\ref{rel:combine}), we do not expect (or need) the new network to perform well right away. Instead, it will be trained for several epochs in the next iteration of PBT-NAS, the same as the other networks in the population. 

%% file: 4_experiments.tex
\section{Experiment setup}
\label{exp}

\subsection{General}

We evaluate PBT-NAS on two tasks known to require careful tuning of network architecture and hyperparameters: GAN training and RL for visual control. In these settings, architecture can strongly influence performance~\cite{gui2021review, sinha2020d2rl}. We consider non-trivial architecture search spaces, see Sections~\ref{exp:setup:gan} and~\ref{exp:setup:rl}. We would like to emphasize that achieving a state-of-the-art result on the chosen tasks is not our goal, instead we aim to demonstrate the feasibility of architecture search via simultaneous training and architecture mixing on tasks where performance strongly depends on architecture.

Hyperparameters of PBT-NAS are population size $N$, step size $e\_step$, selection parameter $\tau$ (we use the default value from PBT, $25\%$, in all experiments), probability $p$ of replacing a layer (which is also set to $25\%$). We aim to avoid unnecessary hyperparameter tuning to see if our approach is robust enough to perform well without it and to save computational resources.

The experiments were run in a distributed way, the details on used hardware and on GPU-hour costs of experiments are given in Appendix~\ref{app:impl}. The algorithms used the amount of compute equivalent to training $N$ networks. Every experiment was run three times, we report the mean and standard deviation of the performance of the best solution from each run. 
We use the Wilcoxon signed-rank test with Bonferroni correction for statistical testing (target $p$-value 0.05, 4 tests, corrected $p$ 0.0125, mentions of statistical significance in the text imply smaller $p$, all $p$-values are reported in Appendix~\ref{app:stats}). Our code is available at \url{https://github.com/AwesomeLemon/PBT-NAS}, it includes configuration files for all experiments.

\subsection{GANs} \label{exp:setup:gan}

In AdversarialNAS~\cite{gao2020adversarialnas}, the authors describe searching for a GAN architecture (for unconditional generation) in a search space where random search achieved poor results~---~this motivated us to adopt this search space, which we refer to as \texttt{Gan}. In AdversarialNAS, both generator and discriminator architectures are searched for but we noticed that in the official implementation, the searched discriminator is discarded, and an architecture from the literature~\cite{gong2019autogan} is used instead. This prompted us to create an extended version of the search space (which we call \texttt{GanHard}) that includes discriminator architectures resembling the one manually selected by the AdversarialNAS authors.
AdversarialNAS cannot be used to search in \texttt{GanHard} because some of the options cannot be searched for via continuous relaxation (one example is searching whether a layer should downsample: since output tensors with and without downsampling have different dimensions, a weighted combination cannot be created). 

Specifics of search spaces are not critical for our research, so we give condensed descriptions here, see Appendix~\ref{app:gan_ss} and our code for more details. 

In \texttt{Gan}, operations for three DARTS-like~\cite{liu2018darts} cells are searched (each cell is a Directed Acyclic Graph (DAG) with operations on the edges; in contrast to DARTS, each cell may have a different architecture). Additionally, inspecting the code of AdversarialNAS showed that the output of some cells is pointwise summed with a projection of a part of a latent vector. Each such projection is a single linear layer mapping a part of a latent vector to a tensor of the same dimensionality as the cell output: $(\#channels, width, height)$. These projections contain many parameters and are therefore an important part of the architecture. In the code of AdversarialNAS, these projections are adjusted for each dataset. As to the discriminator, the architecture from~\cite{gong2019autogan} is used.

Next, we describe \texttt{GanHard}. In \texttt{GanHard}, the parameters of the projections in the generator can be searched for. We additionally treat the layer mapping latent vector to the input of the generator as a projection, since it is conceptually similar. The projections (one per cell) can be enabled or disabled, except for the first one (connected to generator input) which is always enabled. A projection can take as input either the whole latent vector or the corresponding one-third of it (the first third of the vector for the first projection, etc.). 
There are three options for the spatial dimensions of the output of a projection: \emph{target} (equal to the output dimensions of the corresponding cell), \emph{smallest} (equal to the input dimensions of the first cell), and \emph{previous} (equal to the output dimensions of the previous cell).
Since the projection output is summed with the cell output pointwise, the dimensions need to match, which is not the case for the last two options. To upsample the tensor to the target dimensions, either $bilinear$ or $nearest\_neighbour$ interpolation is used, which is also a part of the search space. Finally, given that a projection is a large linear layer with potentially millions of parameters (which makes overfitting plausible), we introduce an option for a dropout layer in the projection, with possible parameters $0.0, 0.1, 0.2$.

The discriminator search space in \texttt{GanHard} is based on the one in AdversarialNAS, the discriminator has 4 cells (each being a DAG with two branches), each cell has a downsampling operation in the end. However, we noticed that the fixed architecture from~\cite{gong2019autogan} that is used for final training of AdversarialNAS downsamples only in the first two cells. Additionally, in the first cell, the input is downsampled rather than the output. We amend the discriminator search space to contain a similar architecture. Firstly, we search whether each cell should downsample or not. Secondly, we add options for downsampling operations that are performed at the start of each branch rather than at the end of them. To enrich the search space further, we add two more nodes to each cell. 

The number of variables in \texttt{Gan} is 21 and the size of the search space is $\approx 3.4 \cdot 10^{19}$. In \texttt{GanHard}, there are 72 variables (32 for the generator, 40 for the discriminator), and the size of the search space is $\approx 2.9 \cdot 10^{53}$.

Following AdversarialNAS, we run the experiments on CIFAR-10~\cite{krizhevsky2009learning} and STL-10~\cite{coates2011analysis}, using both \texttt{Gan} and \texttt{GanHard}. In AdversarialNAS, the networks were trained for 600 epochs. We reduce that number to 300 epochs to save computation time (preliminary experiments showed diminishing returns to longer training), for the other hyperparameters, the same values as in AdversarialNAS are used.

The Frechet Inception Distance (FID)~\cite{heusel2017gans} is a commonly used metric for measuring GAN quality. We use its negation as the objective function, computing it on 5,000 images during the search. For reporting the final result, the FID for the best network is computed on 50,000 images. We additionally report the Inception Score (IS)~\cite{salimans2016improved}, another common metric of GAN quality. The idea behind both FID and IS is to compare representations of real and generated images.

Based on preliminary experiments, the population size $N$ is set to 24, and $e\_step$ is set to 10.

\subsection{RL} \label{exp:setup:rl}

We build upon DrQ-v2~\cite{yarats2022mastering}, a model-free RL algorithm for visual continuous control. DrQ-v2 achieves great results on the Deep Mind Control benchmark~\cite{tassa2018deepmind}, solving many tasks. Searching for architectures for solved tasks is not necessary, therefore for our experiments we chose tasks where DrQ-v2 did not achieve the maximum possible performance: Quadruped Run, Walker Run, Humanoid Run. 

DrQ-v2 is an actor-critic algorithm with three components: \emph{1)} an encoder that creates a representation of the pixel-based environment observation, \emph{2)} an actor that, given the representation, outputs probabilities of actions, and \emph{3)} a critic that, given the representation, estimates the Q-value of the state-action pair (the critic contains two networks because double Q-learning is used). 

We design the search space to include the architectures of all the components of DrQ-v2. Each network has three searchable layers. For the encoder, the options are Identity, Convolution \{3x3, 5x5, 7x7\}, ResNet~\cite{he2016deep} block \{3x3, 5x5, 7x7\}, Separable convolution \{3x3, 5x5, 7x7\}. For the actor and both networks of the critic, the available layers are Identity, Linear, and Residual~\cite{bjorck2021towards} with multiplier $0.5$ or $2.0$. Additionally, we search whether to use Spectral Normalization~\cite{miyato2018spectral} (for each network separately) and which activation function to use in each layer (options: Identity, Tanh, ReLU, Swish). We also search the dimensionality of representation: in DrQ-v2 it was set to either 50 or 100 depending on the task, we have $25, 50, 100,$ and $150$ as options.

There are 36 variables in total, the search space size is $\approx 4.6 \cdot 10^{21}$.

The hyperparameters of DrQ-v2 are used without additional tuning. For the Walker and Humanoid tasks, DrQ-v2 uses a replay buffer of size $10^6$. Our servers do not have enough RAM to allow for such a buffer size when many agents are training in parallel, therefore for these tasks, we use a shared replay buffer (proposed in~\cite{franke2020sample}): different agents can learn from the experiences of each other. To run in a distributed scenario with no shared storage, the buffers are only shared by the networks on the same machine. For fairness, all the baselines also use a shared buffer per machine.

Based on preliminary experiments, the population size $N$ is set to 12. For the Quadruped and the Walker tasks, the DrQ-v2 agent used $3\cdot10^6$ frames. We use the same number of frames per agent, which means that $N$ times more total frames are used. For the Humanoid task, $3\cdot10^7$ frames were used in DrQ-v2, we use only $1.5\cdot10^7$ per agent to save computation time. For uniformness of notation with GANs, we also use "epoch" in the context of RL, one epoch is defined as $10^4$ frames. This means that 300 epochs are used for the Quadruped and Walker tasks, the same as for GANs, and $e\_step$ is also set to 10. Similar to BG-PBT~\cite{wan2022bayesian}, our preliminary experiments showed that having longer periods without selection at the start of the training is beneficial, therefore during the first half of the training, $e\_step$ is doubled from 10 to 20 epochs for the Quadruped and Walker tasks. Since for Humanoid only half the training is performed (in terms of frames per agent), the step size is fixed at 100 epochs (scaled up from 10 proportionally to the increase in the number of frames).

\subsection{Baselines}
We consider two general baselines that parallelize well and that can search in the proposed challenging search spaces.
\begin{enumerate}
\item \textbf{Random search.} $N$ architectures are randomly sampled and trained.

\item \textbf{SEARL-like mutation}. In order to fairly evaluate the performance of a mutation-based architecture search approach like SEARL~\cite{franke2020sample}, we replace the mixing operator of PBT-NAS with the mutation operator from SEARL, adapting it to be applicable to both GAN and RL settings: with equal probability, either \textit{(a)} one variable in the architecture encoding is resampled, \textit{(b)} weights are mutated using the procedure from SEARL, or \textit{(c)} no change is performed.
\end{enumerate}

AdversarialNAS is a specialized baseline only capable of searching in \texttt{Gan}. In~\cite{gao2020adversarialnas}, the performance for only one seed was reported. We run the official implementation with 5 seeds and report the mean and standard deviation of performance.

\section{Results}

\subsection{PBT-NAS vs. the baselines} \label{exp:betterthanbase}

As can be seen in Table~\ref{tab:perf_c10}, PBT-NAS achieves the best performance among all tested approaches in all GAN settings\footnote{When searching in \texttt{Gan} for STL-10, we faced reproducibility issues (despite using the official implementation), see Appendix~\ref{app:stl10} for results and discussion.}. The improvements in FID over both random search and SEARL-based mutation are statistically significant. Despite the claim of~\cite{gao2020adversarialnas} that random search performs poorly in \texttt{Gan}, we find that the gap between it and AdversarialNAS~\cite{gao2020adversarialnas} on CIFAR-10 is small, and the difference between all algorithms is overall not large. The decreased performance of random search in \texttt{GanHard} shows that \texttt{GanHard} is indeed a more challenging search space. The results of searching in this space for CIFAR-10 and STL-10 show a clear improvement of PBT-NAS over the baselines in terms of FID. IS is better in the majority of settings. 

\begin{figure}[h]
\centering
    \subfigure[Quardruped Run]{\includegraphics[width=0.46\linewidth]{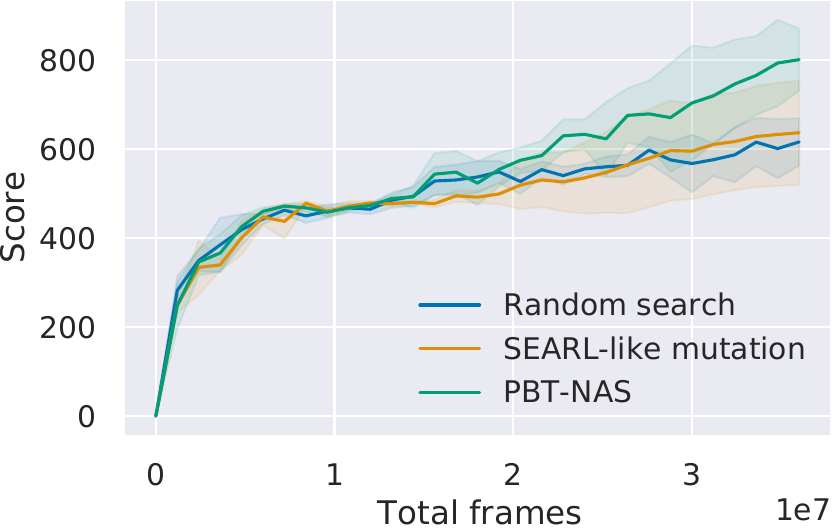}}
    ~
    \hspace{0.1cm}
    ~
    \subfigure[Walker Run]{\includegraphics[width=0.46\linewidth]{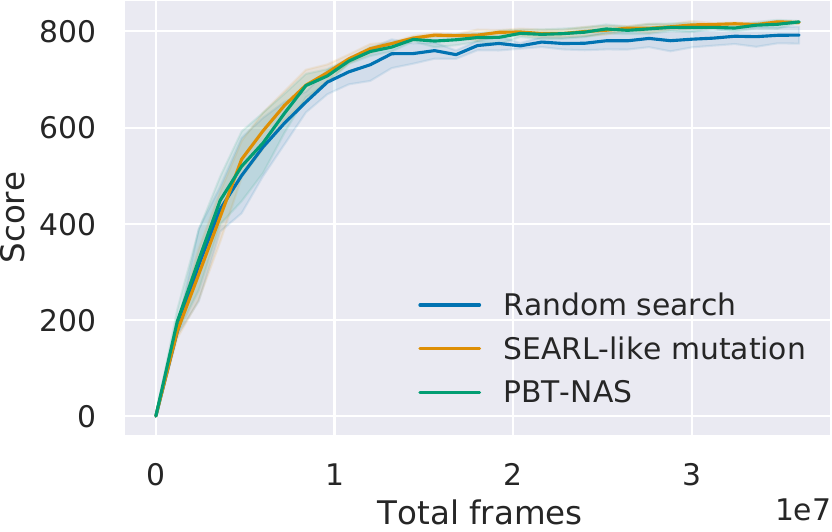}}

    \subfigure[Humanoid Run]{\includegraphics[width=0.46\linewidth]{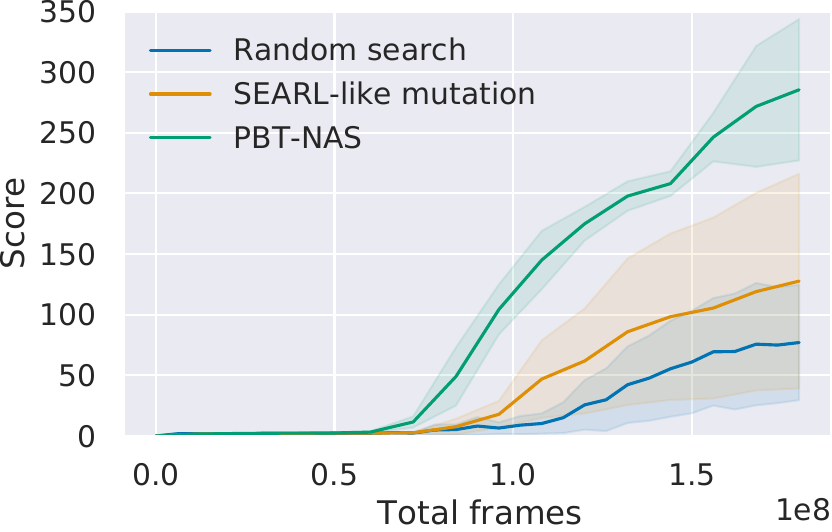}}
\caption{Results for RL tasks, mean $\pm$ st. dev. (shaded area).}
\label{fig:perf_rl}
\end{figure}

PBT-NAS is also the best among alternatives on RL tasks, achieving better anytime performance, as shown in Figure~\ref{fig:perf_rl} (the improvements in score over both random search and SEARL-based mutation are statistically significant). For Walker Run, there is no meaningful difference between algorithms, as the task is solved by all tested approaches, demonstrating that for differences between performance of the algorithms to be clear, both the RL task and the search space need to be of significant complexity.

\begin{table*}[tp]
    \centering
    \caption{Results for GAN training (mean $\pm$ st. dev.). The best value in each column is in bold.}
    \begin{NiceTabular}{@{}lcccc|cc@{}} \toprule
    \Block{3-1}{Algorithm} & \Block{1-4}{CIFAR-10} & & & & \Block{1-2}{STL-10}\\
    \cmidrule{2-7}& \Block{1-2}{\texttt{Gan}} & & \Block{1-2}{\texttt{GanHard}} & & \Block{1-2}{\texttt{GanHard}} \\
    \cmidrule{2-3} \cmidrule{4-5} \cmidrule{6-7}  
    & FID $\downarrow$ & IS $\uparrow$ & FID $\downarrow$ & IS $\uparrow$ & FID $\downarrow$ & IS $\uparrow$ \\
    \midrule
    AdversarialNAS~\cite{gao2020adversarialnas} & $12.29_{\pm 0.80}$ & $8.47_{\pm 0.14}$ & --- & --- &  --- & --- \\
    Random search & $13.39_{\pm 0.28}$ & $8.22_{\pm 0.15}$ & $16.79_{\pm 0.97}$ & $7.80_{\pm 0.14}$ & $28.58_{\pm 1.77}$ & $9.33_{\pm 0.18}$\\
    SEARL-like mutation~\cite{franke2020sample} & $13.78_{\pm 1.02}$ & $8.38_{\pm 0.05}$ & $15.72_{\pm 2.22}$ & $\mathbf{8.26}_{\pm 0.21}$ & $26.94_{\pm 0.93}$ & $9.66_{\pm 0.27}$\\
    PBT-NAS & $\mathbf{12.21}_{\pm 0.16}$ & $\mathbf{8.63}_{\pm 0.17}$  & $\mathbf{13.25}_{\pm 1.64}$ & $8.25_{\pm 0.27}$ & $\mathbf{25.11}_{\pm 0.94}$ & $\mathbf{9.71}_{\pm 0.09}$\\
    \bottomrule
    \end{NiceTabular}
    \label{tab:perf_c10}
\end{table*}

\vspace{-10pt}
\subsection{Mixing networks is better than cloning good networks}

In order to show that creating new architectures makes a difference, we run a "No mixing" ablation: every component of PBT-NAS is kept the same, except that a new model is created by mixing a well-performing model with itself (rather than with another well-performing model). This way, no new architecture is produced, but the other benefits of PBT-NAS remain (e.g., replacing poorly-performing networks with well-performing ones). As seen in Table~\ref{tab:ablation}, this degrades the performance, clearly showing the impact that creating a better architecture can have.

\begin{table}
    \centering
    \caption{Results of ablation studies (mean $\pm$ st. dev.). The best value in each column is in bold.}
    \begin{NiceTabular}{@{}lcc@{}} \toprule
    \Block{3-1}{Algorithm} & FID $\downarrow$ & Score $\uparrow$ \\
    
    & (CIFAR-10, & (Quadruped Run) \\
    & \texttt{GanHard}) & \\
    \midrule
    PBT-NAS (default) & $\mathbf{13.25}_{\pm 1.64}$ & $\mathbf{801}_{\pm 70}$ \\
    No mixing & $14.90_{\pm 1.29}$ & $672_{\pm 53}$\\
    Shrink-perturb coefficients: & &\\
    \hspace{0.5cm} [1, 0] --- copy exactly & $14.94_{\pm 0.56}$ & $699_{\pm 4}$\\
    \hspace{0.5cm} [0, 1] --- reinitialize randomly & $15.06_{\pm 2.59}$ & $532_{\pm 62}$\\
    \bottomrule
    \end{NiceTabular}
    \label{tab:ablation}
\end{table}

\vspace{-10pt}
\subsection{Shrink-perturb is the superior way of weight inheritance} \label{abl:shpe_crucial}

Table~\ref{tab:ablation} shows that copying weights from the donor without change (shrink-perturb parameters $[1, 0]$), or replacing them with random weights (shrink perturb $[0, 1]$) leads to worse results in comparison to the usage of shrink-perturb. Thus, in our settings, using shrink-perturb is the best method to inherit the weights. The default parameters  of shrink-perturb from~\cite{Zaidi_Berariu_Kim_Bornschein_Clopath_Teh_Pascanu_2022} ($[0.4, 0.1]$) worked well in PBT-NAS without any tuning.

In~\cite{Zaidi_Berariu_Kim_Bornschein_Clopath_Teh_Pascanu_2022}, shrink-perturb was found to benefit performance, thus raising the question if using it gives PBT-NAS an unfair advantage that is not related to NAS. In order to test this, we added shrink-perturb to random search. As shown in Table~\ref{tab:shpe_random_search}, performance deteriorates, indicating that using shrink-perturb with default parameters in our setting is not helpful outside the context of NAS.

\begin{table}
    \centering
    \caption{The effect of using shrink-perturb in random search}
    \begin{NiceTabular}{@{}lcc@{}} \toprule
    Use shrink-perturb & FID $\downarrow$ & Score $\uparrow$ \\
    in random search & (CIFAR-10, \texttt{GanHard}) & (Quadruped Run) \\
    \midrule
    No (default) & $16.79_{\pm 0.97}$ & $616_{\pm53}$ \\
    Yes & $22.39_{\pm 2.64}$ & $498_{\pm30}$ \\
    
    \bottomrule
    \end{NiceTabular}
    \label{tab:shpe_random_search}
\end{table}

\vspace{-10pt}
\subsection{Increasing population size improves performance} \label{abl:scaling}

\begin{figure}[h]
\centering
    \subfigure[CIFAR-10, \texttt{GanHard} (FID $\downarrow$)]{\includegraphics[width=0.43\linewidth]{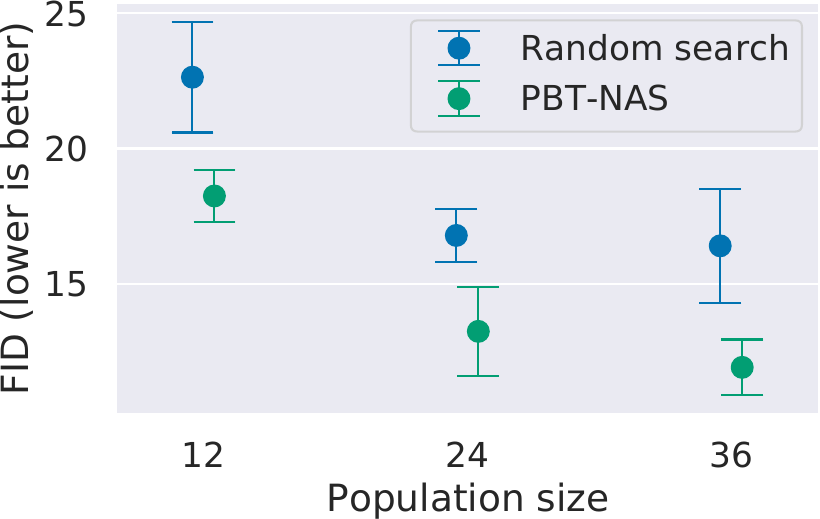}}
    ~
    \hspace{0.1cm}
    ~
    \subfigure[Quadruped Run (Score $\uparrow$)]{\includegraphics[width=0.43\linewidth]{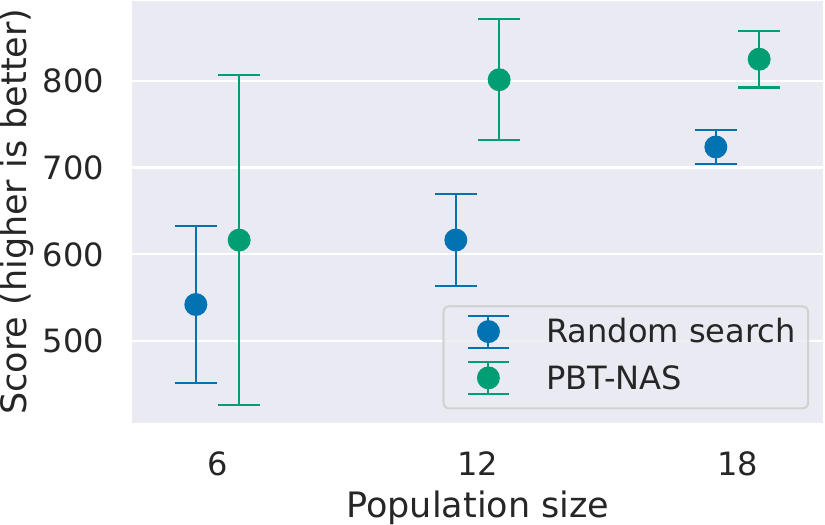}}
\caption{Impact of scaling population size, mean $\pm$ st. dev.}
\label{fig:scaling}
\end{figure}

We design our algorithm to be highly parallel and scalable. Figure~\ref{fig:scaling} demonstrates that as the population size increases, the performance strictly improves (although diminishing returns can be observed). Given enough GPUs, the increased population size will not meaningfully increase wall-clock time, since every population member can be evaluated in parallel.

\subsection{Model soups}

As mentioned in Section~\ref{rel:combine}, the idea of a model soup~\cite{wortsman2022model} is to improve performance by averaging weights of closely-related neural networks. As such, it seems like an especially good fit for the PBT setting: although the networks in the population start from different weights (and different architectures in the case of PBT-NAS), as worse networks are replaced by offspring of better networks, the population gradually converges. Since creating a model soup is done after training and requires a negligible amount of computation (evaluating at most $N$ models), its inclusion into PBT-like algorithms could give an almost free performance improvement. Therefore, we create model soups following the greedy algorithm from~\cite{wortsman2022model}.

Table~\ref{tab:soup_gan} shows that soups improve GAN FID by approximately 0.4 points. For RL, however, there is no improvement when both the encoder and the actor are averaged (Table~\ref{tab:soup_rl}). We hypothesize that different actors may have dissimilar internal representations implementing different behaviour logic, unlike the encoders that only convert pixel inputs into representations. Therefore, we tried to separately average encoders, or actors. The results with averaged encoders are the best overall but they still do not lead to improved performance. For Walker Run, the task where performance is saturated, there is no difference between settings.

\begin{table}
    \centering
    \caption{The difference in metrics between a model soup and the best individual model (GAN), mean $\pm$ st. dev.}
    \begin{NiceTabular}{@{}lcc@{}} \toprule
    \Block{2-1}{Dataset} & \Block{1-2}{\texttt{GanHard}}\\
    \cmidrule{2-3} & $\Delta$ FID $\downarrow$ &  $\Delta$ IS $\uparrow$\\
    \midrule
    CIFAR-10 & $-0.48_{\pm 0.34}$ & $0.07_{\pm 0.05}$\\
    STL-10 &  $-0.35_{\pm 0.26}$ & $0.04_{\pm 0.25}$ \\
    \bottomrule
    \end{NiceTabular}
    \label{tab:soup_gan}
\end{table}

\begin{table}
    \centering
    \caption{The difference in score between a model soup and the best individual model (RL), mean $\pm$ st. dev.}
    \begin{NiceTabular}{@{}lccc@{}} \toprule
    \Block{2-1}{What to average} & \Block{1-3}{$\Delta$ Score $\uparrow$}\\
    \cmidrule{2-4} & Quadruped & Walker & Humanoid \\
    \midrule
    Encoder & $-6_{\pm 10}$ & $-1_{\pm 4}$& $-23_{\pm 19}$\\
    Actor & $-196_{\pm 275}$ & $-4_{\pm 7}$ & $-244_{\pm 32}$\\
    Both & $-142_{\pm 193}$ & $0_{\pm 6}$& $-223_{\pm 20}$ \\
    \bottomrule
    \end{NiceTabular}
    \label{tab:soup_rl}
\end{table}

Previously, soups were only demonstrated for classification tasks, so it is interesting to see that they could also be beneficial in GANs. While no improvement was seen for RL, the fact that only the vision-related network, the encoder, could be averaged without large performance degradation hints at the limitations of the technique.

%% file: 5_discussion.tex
\section{Discussion}
\label{discussion}

We have introduced PBT-NAS, a NAS algorithm that creates new architectures by simultaneously training and mixing a population of neural networks. PBT-NAS brings the efficiency of PBT (designed for hyperparameter optimization) to NAS, providing a novel way to search for architectures. As computation power grows, especially in the form of multiple affordable GPUs, having parallelizable and scalable algorithms such as PBT-NAS becomes more important. At the same time, this computation power is not limitless, and reusing the partly-trained weights during architecture search is important from the perspective of search efficiency. 

Currently, a large amount of effort in single-objective NAS research is directed at searching classifier architectures in cell-based search spaces, which are quite restrictive, and where random search achieves competitive results~\cite{li2020random, yangevaluation}. We think that pivoting to more challenging search spaces and tasks could lead to NAS having a larger impact (e.g., in constructing state-of-the-art architectures, which is still mostly done by hand), and to comparisons between NAS algorithms leading to clearer differences. In Section~\ref{exp:betterthanbase}, we showed that PBT-NAS could search in the challenging \texttt{GanHard} space, where an existing efficient algorithm, AdversarialNAS, could not be applied.

One limitation of exchanging layers during training is the requirement that different layer options (in the same position) need to be interoperable: the  activation tensors they produce should be possible for the next layer to take as input (so that after replacing a layer, the architecture remains valid). This means that the number of neurons can be searched only when it does not influence the output shape.
This could be addressed by e.g., duplicating neurons if there are too few of them and removing excessive ones if there are too many. Another limitation arises due to the greedy nature of PBT-NAS: architectures are selected based on their intermediate performance, and, therefore, suboptimal architectures can be selected when early performance of an architecture is not representative of the final one.

Achieving good performance in different tasks with minimal hyperparameter tuning is a desirable property for a NAS algorithm. We used hyperparameters from the literature without tuning both in GAN training and in RL, as well as relying on default selection strategy from PBT. PBT-NAS outperformed baselines despite using these default values, tuning them could potentially further improve the results.

%% file: 6_conclusion.tex
\vspace{-5pt}
\section{Conclusion}
\label{conclusion}

In this paper we designed and evaluated PBT-NAS, a novel way to search for an architecture by mixing different architectures while they are being trained. 
We find that adapting the weights with the shrink-perturb technique during mixing is advantageous compared to copying or randomly reinitializing them.

PBT-NAS is shown to be effective on challenging tasks (GAN training, RL), where it outperformed considered baselines. At the same time, it is efficient, requiring training of only tens of networks to explore large search spaces. The algorithm is straightforward, parallelizes and scales well, and has few hyperparameters.

While in this work only NAS was considered, in the future, PBT-NAS could be adapted to simultaneously search for hyperparameters of neural network training, and of the algorithm itself, both of which would be necessary in order to fully automate the process of neural network training.

%% file: 7_appendix.tex
\appendix

\section{Visualizing search progress} \label{app:viz}

In order to visualize the search process, we track the origin of all the layers in the population. Initially, the layers of the $i$-th population member are specified to have origin $i$. When a new network is created by mixing different networks, its layers will have different origins. Over time, as worse-performing networks are replaced with the results of mixing better-performing ones, the successful layers constitute an increasing proportion of all layers in the population.

\begin{figure}[h]
\centering
    \subfigure[Score 910]
    {\includegraphics[width=0.47\linewidth]{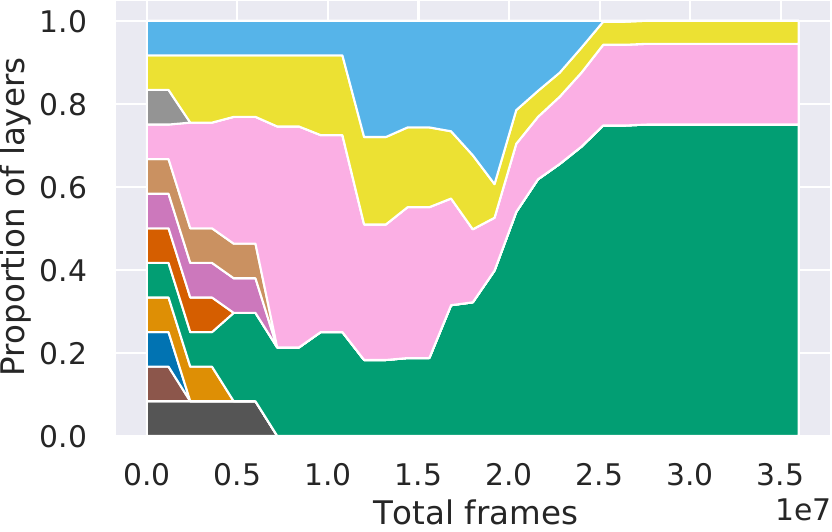}\label{fig:vis_progress:a}}
    ~
    \hspace{0.01cm}
    ~
    \subfigure[Score 725]
    {\includegraphics[width=0.47\linewidth]{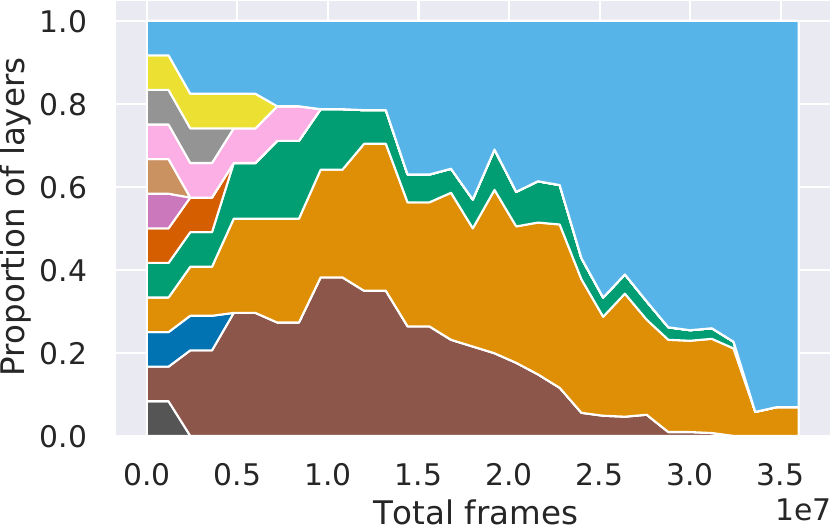}\label{fig:vis_progress:b}}
     ~
    \subfigure
    {\includegraphics[width=0.9\linewidth]{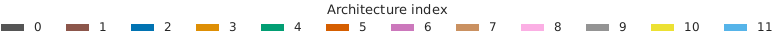}}
\caption{Proportion of layers from different initial architectures in the population over time (Quadruped Run). The experiments with the best \protect{\subref{fig:vis_progress:a}} and the worst \protect{\subref{fig:vis_progress:b}} scores are shown.
}
\label{fig:vis_progress}
\end{figure}

We visualize the experiments on the Quadruped Run task that achieved the best (Figure~\ref{fig:vis_progress:a}) and the worst (Figure~\ref{fig:vis_progress:b}) scores (out of the three seeds).
In the experiment in Figure~\ref{fig:vis_progress:a}, many architectures are successfully mixed, with the final population containing layers coming from several initial architectures. On the contrary, in the experiment in Figure~\ref{fig:vis_progress:b}, one architecture largely takes over, with a small part inherited from a second architecture. 

\section{Additional Bayesian Optimization baseline} \label{app:bohb}
In this section we compare PBT-NAS to an additional baseline, Bayesian Optimization Hyperband (BOHB)~\cite{falkner2018bohb}. BOHB is a well-established, efficient, and parallel Bayesian Optimization (BO) algorithm designed for hyperparameter optimization and NAS.

BO algorithms can be applied to almost any NAS search space, but typically have limited parallelization capability, and can struggle in high-dimensional spaces, especially if there are too few evaluations available. Our experiments in the main text were conducted in exactly such a scenario: the search spaces were high-dimensional (the \texttt{GanHard} space has 72 variables, the RL space has 36), and the number of total evaluations is smaller than the number of variables (24 evaluations for \texttt{GanHard}, 12 for RL). In addition, PBT-NAS and the baselines considered in the main text are fully parallelizable (unlike BO algorithms).

We report the performance of BOHB in two settings (in which all our ablations were done). BOHB is run with the same budget as PBT-NAS both for Reinforcement Learning (the Quadruped Run task, total number of epochs equal to fully training 12 architectures) and GAN training (\texttt{GanHard} space, CIFAR-10 dataset, total number of epochs equal to fully training 24 architectures). Same as in the main text, every experiment is repeated three times, we report the mean and standard deviation of the performance of the best architecture.

\begin{table}
    \centering
    \caption{Results of running BOHB. The best value in each column is in bold.}
    \begin{NiceTabular}{@{}lcc@{}} \toprule
    Algorithm & FID $\downarrow$ & Score $\uparrow$ \\
     & (CIFAR-10, \texttt{GanHard}) & (Quadruped Run) \\
    \midrule
    PBT-NAS & $\mathbf{13.25}_{\pm 1.64}$ & $\mathbf{801}_{\pm 70}$ \\
    BOHB & $18.95_{\pm 1.48}$ & $621_{\pm28}$ \\
    
    \bottomrule
    \end{NiceTabular}
    \label{tab:bohb}
\end{table}

As can be seen in Table~\ref{tab:bohb}, BOHB  is outperformed by PBT-NAS in both settings. The BOHB results for GAN training are particularly poor, we speculate that BOHB suffers from excessive greediness in this setting (it early-stops two-thirds of solutions at each fidelity). PBT-NAS, while also greedy, has lower selection pressure, so it is affected less.

\section{Statistical testing} \label{app:stats}

Table~\ref{tab:pvalues} lists p-values for one-sided Wilcoxon pairwise rank tests~\cite{wilcoxon1992individual} with Bonferroni correction~\cite{dunn1961multiple}. In the GAN setting, the null hypothesis is that Algorithm 1 has a higher FID than Algorithm 2. In the RL setting the null hypothesis is that Algorithm 1 has a lower score than Algorithm 2.

In the GAN setting, results for CIFAR-10 using \texttt{Gan}, \texttt{GanHard}, and STL-10 using \texttt{GanHard} are tested together to increase sample size (total sample size is 9). Similarly to increase sample size, in the RL setting all the tasks (Quadruped Run, Walker Run, Humanoid Run) are tested together (total sample size is 9).
P-values below the significance threshold of 0.0125 are highlighted (target $p$-value=0.05, 4 tests, corrected $p$=0.0125). 

\begin{table}[htbp]
\centering
  \caption{P-values of conducted experiments.}
  \label{tab:pvalues}
  \begin{tabular}{cccl}
    \toprule
    Algorithm 1 & Algorithm 2 & Setting & P-value\\
    \midrule
    PBT-NAS & Random search & GAN & \cellcolor{mycolor} 0.001953125 \\
    PBT-NAS & SEARL-like mutation & GAN & \cellcolor{mycolor} 0.00390625 \\
    PBT-NAS & Random search & RL & \cellcolor{mycolor} 0.001953125 \\
    PBT-NAS & SEARL-like mutation & RL & \cellcolor{mycolor} 0.005859375 \\
  \bottomrule
\end{tabular}
\end{table}

\section{Reproducibility issues with \texttt{Gan} on STL-10} \label{app:stl10}

In~\cite{gao2020adversarialnas}, results for STL-10 are reported on a single seed, we ran the experiment 5 times using the official implementation. The resulting FID for STL-10 increases by about 10 points, which is a substantial drop in performance (see Table~\ref{tab:perf_stl_gan}). The immediate cause is the divergence of the training procedure before good results could be achieved. However, as the official implementation was used, it is unclear why this divergence appeared consistently for all seeds and why none of the five seeds achieved the performance reported in~\cite{gao2020adversarialnas}. The authors did not respond when we notified them of the problem. We are also aware of an independent reproduction attempt running into the same issue.

Since our code relies on the AdversarialNAS codebase, whatever the issue is, it influenced all our experiments with \texttt{Gan} on STL-10, bringing their validity into question. Nonetheless, for transparency, our results (for three seeds) are reported in Table~\ref{tab:perf_stl_gan}. PBT-NAS achieves the best performance out of the approaches tested by us.

Note that we did not face the reproducibility issue with \texttt{Gan} on CIFAR-10, as the reproduced result (FID $12.29_{\pm 0.80}$) was close to the reported one (FID $10.87$~\cite{gao2020adversarialnas}). Also of note is that with \texttt{GanHard} on STL-10, results were even better than in~\cite{gao2020adversarialnas} (see Table 1 in the main text): FID of $25.11_{\pm 0.94}$. Since the same training code was used for \texttt{Gan} and \texttt{GanHard}, this implies that architectures could be the root cause of the problem, with architectures from \texttt{Gan} being a poor fit for STL-10, in contrast to architectures from \texttt{GanHard}. This is not consistent with the results from the main text for CIFAR-10, where, controlled for the amount of computational effort, better architectures could be found in \texttt{Gan} than in \texttt{GanHard}.

We would also like to note that \texttt{GanHard} was designed before any experiments with STL-10 were run, precluding the possibility that better STL-10 results with \texttt{GanHard} (compared to \texttt{Gan}) were achieved by deliberate search space adaptation to the dataset. STL-10 was our test dataset, the results for which did not influence any design or hyperparameter choices made in the paper.

\begin{table}[tp]
    \centering
    \caption{Additional results for GAN training (mean $\pm$ st. dev.).}
    \begin{NiceTabular}{@{}lcc@{}} \toprule
    \Block{3-1}{Algorithm} & \Block{1-2}{STL-10} & \\
    \cmidrule{2-3} & \Block{1-2}{\texttt{Gan}} & \\
    \cmidrule{2-3}
    & FID $\downarrow$ & IS $\uparrow$ \\
    \midrule
    AdversarialNAS (reported in~\cite{gao2020adversarialnas}) & $26.98$ & $9.63$ \\
    AdversarialNAS (reproduced) & $36.87_{\pm 3.62}$ & $8.90_{\pm 0.32}$ \\
    Random search & $32.54_{\pm 3.20}$ & $9.15_{\pm 0.08}$ \\
    SEARL-like mutation~\cite{franke2020sample} & $33.98_{\pm 4.36}$ & $8.98_{\pm 0.19}$ \\
    PBT-NAS & $29.51_{\pm 0.91}$ & $9.19_{\pm 0.05}$\\
    \bottomrule
    \end{NiceTabular}
    \label{tab:perf_stl_gan}
\end{table}

\section{Illustrations of GAN search spaces} \label{app:gan_ss}

\begin{figure}[ht]
\centerline{\includegraphics[width=0.95\linewidth]{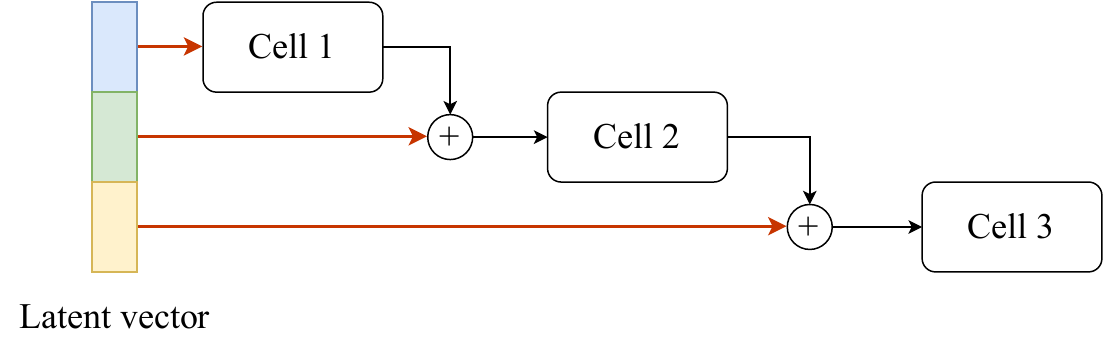}}
\caption{Projections from parts of a latent vector in the macro architecture of AdversarialNAS. Each projection (dark orange) is a linear layer, the output of which is reshaped and pointwise summed with the output of the corresponding cell (except for the first projection, the output of which is the input of the first cell).}
\label{fig:gan_projections}
\end{figure}

\begin{figure}[h]
\centering
    \subfigure[\texttt{Gan}]{\includegraphics[width=0.43\linewidth]{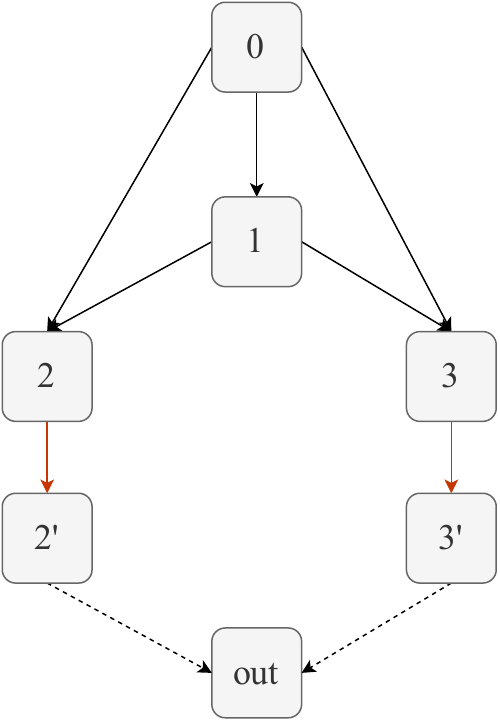}}
    ~
    \hspace{0.1cm}
    ~
    \subfigure[\texttt{GanHard}]{\includegraphics[width=0.43\linewidth]{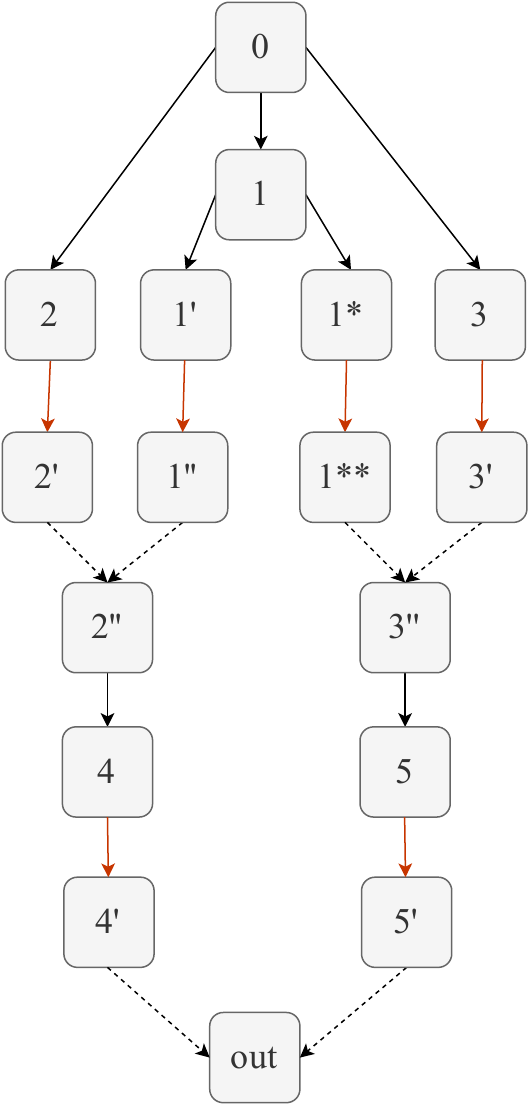}}
    
\caption{Discriminator cell search spaces. Black arrows correspond to normal operations (same as in~\cite{gao2020adversarialnas}, possible operations for the edges are None, Identity, Convolution 1x1 with Dilation=1, Convolution \{3x3, 5x5\} with Dilation=\{1, 2\}), dark orange arrows represent downsampling operations (Average Pooling, Max Pooling, Convolution \{3x3, 5x5\} with Dilation=\{1, 2\}; only parameterless operations (Average Pooling, Max Pooling) are allowed at the start of the branch because they will be applied to several inputs), dashed arrows represent identity. All inputs to a node are summed. For \texttt{GanHard}, downsampling will be either at the start or at the end of a branch (or disabled entirely).}
\label{fig:discr_search_space}
\end{figure}

\section{Implementation details} \label{app:impl}

The experiments were run on servers equipped with three Nvidia A5000 GPUs each. Each server is equipped with 2 Intel(R) Xeon(R) Bronze 3206R CPUs, and 96 GB of RAM. The used OS is Fedora Linux 36. The Ray~\cite{moritz2018ray} framework was used to run the experiments in a distributed fashion. The configuration files specifying versions of all software libraries are included in the source code.

The experiment costs in total GPU hours vary by the setting (CIFAR-10|\texttt{Gan}: 200, CIFAR-10|\texttt{GanHard}: 270, STL-10|\texttt{Gan}: 720, STL-10|\texttt{GanHard}: 1200, Quadruped Run: 210, Walker Run: 330, Humanoid Run: 1300). 